\documentclass[10pt,twocolumn,letterpaper]{article}

\usepackage{iccv}
\usepackage{times}
\usepackage{epsfig}
\usepackage{graphicx}
\usepackage{amsmath}
\usepackage{amssymb}
\usepackage{multirow}
\usepackage{subfigure}
\usepackage{algorithm}
\usepackage{algorithmic}

\usepackage[breaklinks=true,bookmarks=false]{hyperref}

\iccvfinalcopy % *** Uncomment this line for the final submission

\def\iccvPaperID{****} % *** Enter the ICCV Paper ID here

\begin{document}

%%%%%%%%% TITLE
\title{Supervised Online Hashing via Similarity Distribution Learning}

\author{Mingbao Lin$^{1}$, Rongrong Ji$^{1,2,}\thanks{Corresponding Author.}$\,, Shen Chen$^{1}$, Feng Zheng$^{3}$, Xiaoshuai Sun$^{1,2}$, \\ Baochang Zhang$^{4}$, Liujuan Cao$^{1}$, Guodong Guo$^{5}$, Feiyue Huang$^{6}$\\
$^1$Fujian Key Laboratory of Sensing and Computing for Smart City, \\ School of Information Science and Engineering, Xiamen University, China 
\\ $^2$ Peng Cheng Laboratory, China 
\\ $^3$ Department of Computer Science and Engineering, \\ Southern University of Science and Technology, Shenzhen, China.
\\ $^4$ School of Automation Science and Electrical Engineering, Beihang University, China
\\ $^5$ Institute of Deep Learning, Baidu Research
\\ $^6$Tencent Youtu Lab, Tencent Technology (Shanghai) Co., Ltd, China 
\\ \{lmbxmu, chenshen\}@stu.xmu.edu.cn, \{rrji, caoliujuan\}@xmu.edu.cn, zhengf@sustech.edu.cn, \\ xiaoshuaisun.hit@gmail.com, bczhang@buaa.edu.cn, guoguodong01@baidu.com, garyhuang@tencent.com}

\maketitle
%\thispagestyle{empty}

%%%%%%%%% ABSTRACT
\begin{abstract}
Online hashing has attracted extensive research attention when facing streaming data.
Most online hashing methods, learning binary codes based on pairwise similarities of training instances, fail to capture the semantic relationship, and suffer from a poor generalization in large-scale applications due to large variations.
In this paper,
we propose to model the similarity distributions between the input data and the hashing codes, upon which a novel supervised online hashing method, dubbed as Similarity Distribution based Online Hashing (SDOH), is proposed, to keep the intrinsic semantic relationship in the produced Hamming space.
Specifically, we first transform the discrete similarity matrix into a probability matrix via a Gaussian-based normalization to address the extremely imbalanced distribution issue.
And then, we introduce a scaling Student $t$-distribution to solve the challenging initialization problem, and efficiently bridge the gap between the known and unknown distributions.
Lastly, we align the two distributions via minimizing the Kullback-Leibler divergence (KL-diverence) with stochastic gradient descent (SGD), by which an intuitive similarity constraint is imposed to update hashing model on the new streaming data with a powerful generalizing ability to the past data.
Extensive experiments on three widely-used benchmarks validate the superiority of the proposed SDOH over the state-of-the-art methods in the online retrieval task.
\end{abstract}

%%%%%%%%% BODY TEXT
\section{Introduction \label{introduction}}
Hashing based visual search has attracted extensive research attention in recent years due to the rapid growth of visual data on the Internet \cite{gionis1999similarity,weiss2009spectral,gong2011iterative,liu2012supervised,huang2013online,jiang2015scalable,shen2015learning,wang2016learning,liu2018dense,Liu2018Modality,liu2018fast}.
In various scenarios, online hashing has become a hot topic due to the emergence of handling the streaming data, which aims to resolve an online retrieval task by updating the hash functions from sequentially arriving data instances.
On one hand, online hashing takes advantages of traditional offline hashing methods, \emph{i.e.}, low storage cost and efficiency of pairwise distance computation in the Hamming space.
On the other hand, it also merits in training efficiency and scalability for large-scale applications, since the hash functions are updated instantly and solely based on the current streaming data, which is superior to traditional hashing methods based on a hashing model entirely trained from scratch.

%%%%%%%%%%%%%%%%%%%%%%%%%%%%%%%%%%%%%%%%%
\begin{figure}[!t]
\begin{center}
\includegraphics[height=0.33\linewidth]{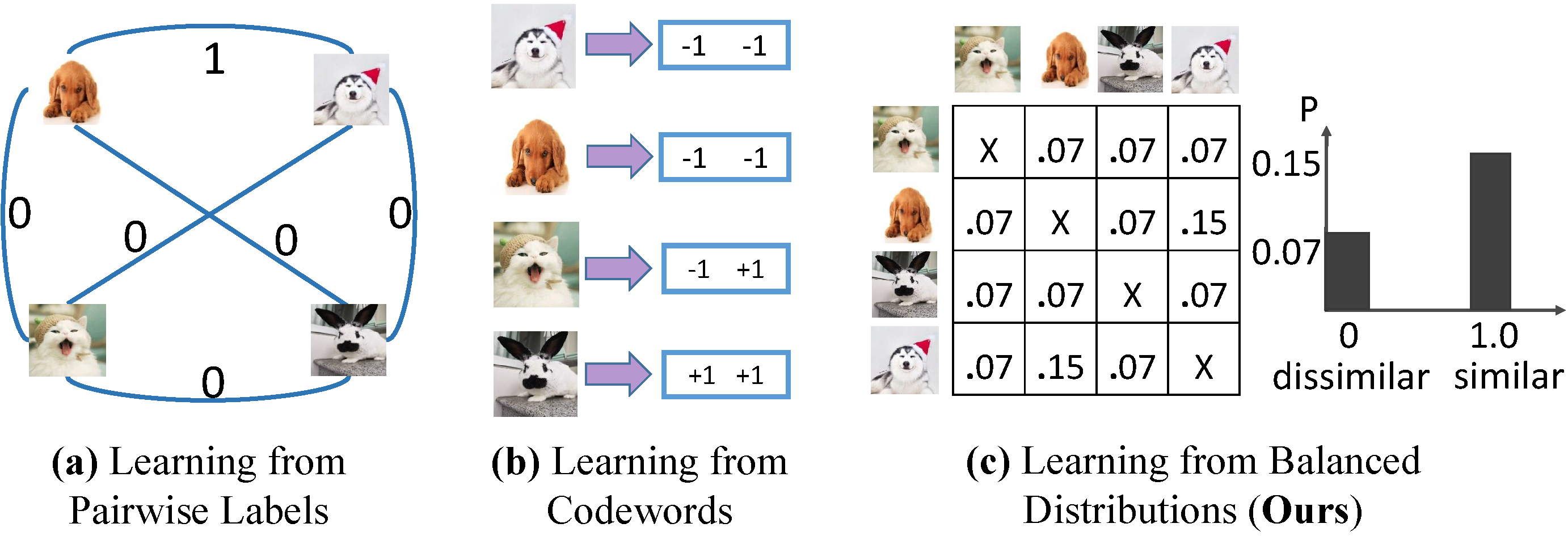}
\end{center}
\vspace{-1em}
\caption{\label{illustration}
Given samples from three classes, \emph{i.e.}, dog, rabbit and cat,
the first kind (a) of supervised online hashing only considers the pairwise similarities, which suffers from a poor generalization.
The second kind (b) assigns a ``codeword" to samples with the same class label, which heavily relies on the quality of codewords.
The proposed SDOH aligns the distributions between the input data and the hashing space (c), which preserves similarity better in the produced Hamming space \cite{lin2010spec,goodfellow2014generative,lin2015semantics,hao2017unsupervised,wu2019deep}.
Different from the previous method \cite{lin2015semantics}, we present two effective and generalized distributions to solve both the imbalanced distribution and poor initialization problems.
}
\vspace{-1.8em}
\end{figure}
%%%%%%%%%%%%%%%%%%%%%%%%%%%%%%%%%%%%%%%%%%%%%%%

%%%%%%%%%%%%%%%%%%%%%%%%%%%%%%%%%%%%%%%%%
\begin{figure*}[!t]
\begin{center}
\includegraphics[height=0.43\linewidth]{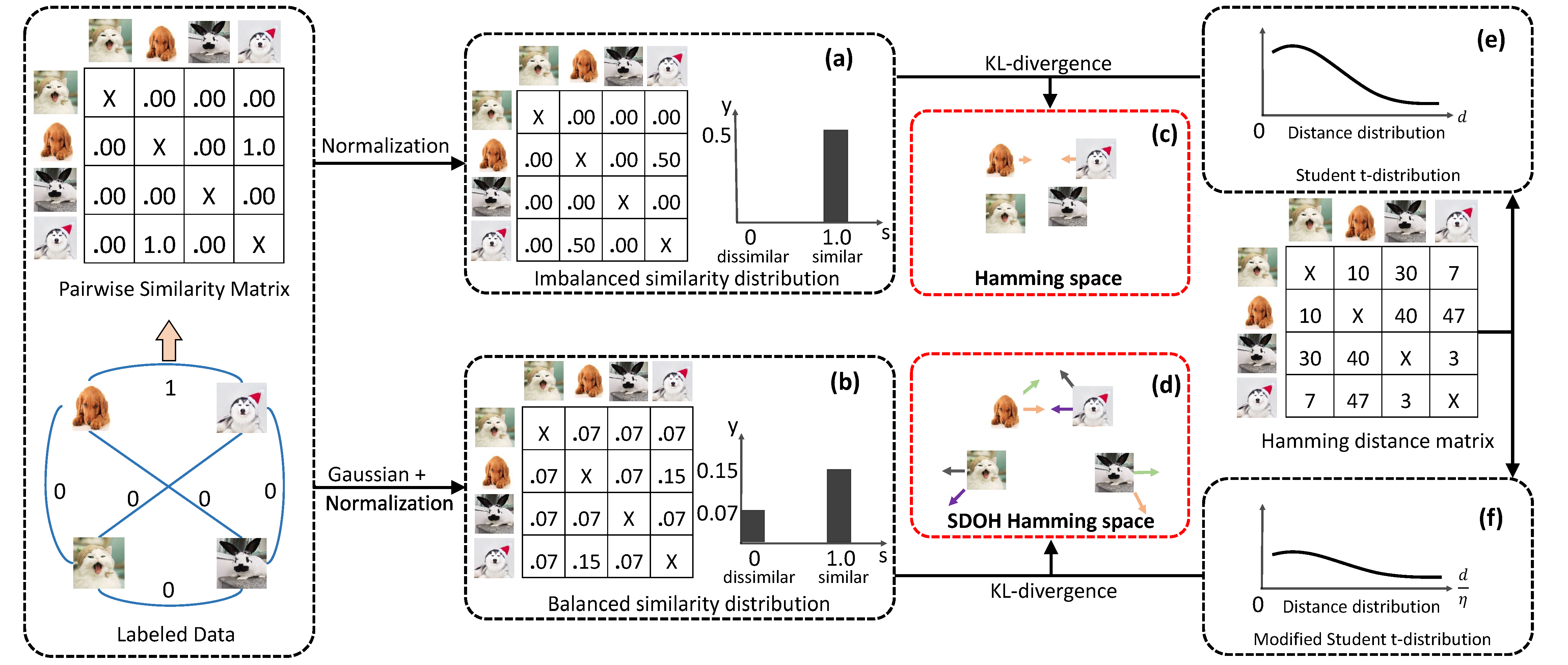}
\end{center}
\vspace{-1em}
\caption{\label{system}
Framework of the proposed SDOH. In the top row, existing methods \cite{lin2015semantics,wu2019deep} construct the similarity distribution (a) and distance distribution in the Hamming space (e) simply via normalization and Student $t$-distribution. Due to the existence of extremely imbalanced distribution in (a) and poor initialization in (e), the learning of KL-divergence heavily relies on similar pairs and cannot converge these two distributions well, leading to a performance degradation (c). The bottom row shows the proposed SDOH. We apply the Gaussian distribution to the similarity matrix before normalization, which relieves the imbalanced distribution problem (b). Further, we introduce a scaling Student $t$-distribution as elaborated in Sec.\,\ref{scheme} to relieve the poor initialization problem (f). Through these operations, the similarity distribution can be well approximated by the learned distance distribution in the Hamming space, where the similarity constraint is preserved (d). As demonstrated in the experiments, the learned hash functions not only capture the pairwise similarities at current stage but also preserve the relationships among different stages.
}
\vspace{-1.6em}
\end{figure*}
%%%%%%%%%%%%%%%%%%%%%%%%%%%%%%%%%%%%%%%%%%%%%%%

%
Several recent endeavors have been made for robust and efficient online hashing, \emph{i.e.}, OKH \cite{huang2013online}, SketchHash \cite{leng2015online}, AdaptHash \cite{cakir2015adaptive}, OSH \cite{cakir2017online}, FROSH \cite{Chen2017FROSHFO}, MIHash \cite{fatih2017mihash}, HCOH \cite{lin2018supervised,lin2019hadamard} and BSODH \cite{lin2019towards}.
Unsupervised online hashing methods, \emph{e.g.}, SketchHash \cite{leng2015online} and FROSH \cite{Chen2017FROSHFO}, consider a sketch of the whole streaming data, which is efficient but lacks in accuracy, as the label information is ignored.

Recent advances have advocated more on supervised online hashing, which yields better results in practice.
As shown in Fig.\,\ref{illustration}(a), early works such as OKH \cite{huang2013online}, AdaptHash \cite{cakir2015adaptive}, MIHash \cite{fatih2017mihash} and BSODH \cite{lin2019towards} utilize label information to define the pairwise similarities between different training instances to guide the learning of hash functions.
However, these methods suffer from a poor generalization. To explain, as demonstrated in a previous work \cite{lin2018supervised}, only pairwise relationships of sequential data at current stage are considered, which ignores the data variations in different stages. As a result, the property of the past data becomes less conspicuous as the dataset grows.
In terms of OSH \cite{cakir2017online} and HCOH \cite{lin2018supervised,lin2019hadamard}, the label information is used to assign ``codeword" from a pre-defined ECOC codebook, as shown in Fig.\,\ref{illustration}(b). And the hash functions map the to-be-learned hashing codes to the assigned ``codeword", which however heavily depends on the quality of ECOC codebook.
Though a recent work in \cite{lin2018supervised} considers the Hadamard matrix \cite{horadam2012hadamard} as the ECOC codebook, it restricts the length of hashing bits to be consistent with the size of the Hadamard matrix.

Despite the extensive progress made, supervised online hashing remains unsolved due to the defect in modeling the supervised cues.
Existing methods only preserve the information from the current data, and their update does not take the distributions of previous data into account.
We argue that, these defects can be compensated by aligning the distributions between the input data and the hashing space when updating, which has been demonstrated informatively beyond online hashing as revealed in \cite{lin2010spec,goodfellow2014generative,lin2015semantics,hao2017unsupervised,wu2019deep}.
Inspired from it, we aim to impose an intuitive constraint on similarity preservation in the Hamming space to capture not only the pairwise similarity at the current stage, but also the semantic relationship among different stages.
By doing so, the learning can take both the information from the current streaming data, but also the past data into account.

In this paper, we propose a novel online hashing method, called Similarity Distribution based Online Hashing (SDOH), which exploits the distribution over different pairwise similarities towards optimal supervised online hashing, as shown in Fig.\,\ref{illustration}(c).
To this end, we first transfer the discrete similarity matrix into a probability matrix via a Gaussian-based normalization.
Noticeably, Lin \emph{et al.} \cite{lin2015semantics} adopted a similar idea which simply obtains a probability matrix via normalization. However, such a normalization may generate an extremely imbalanced distribution (as illustrated in Fig.\,\ref{system}(a)) when facing a highly sparse pairwise discrete similarity matrix.
And the optimization takes a risk of losing the information of dissimilar pairs (see Fig.\,\ref{system}(c)) \cite{arjovsky2017towards}.
Therefore, we introduce a Gaussian distribution to smooth the imbalanced distribution before normalization, which bridges the gap between similar and dissimilar probabilities (as in Fig.\,\ref{system}(b)).
Second, we develop a scaling Student $t$-distribution to transform pairwise distances in the Hamming space into a probability (see Fig.\,\ref{system}(f)).
Different from traditional Student $t$-distribution that suffers from poor performance due to the instability of parameter initialization (see Fig.\,\ref{system}(e)),
the proposed scaling Student $t$-distribution not only improves the performance but also accelerates the training speed (see Fig.\,\ref{system}(d)).
Lastly, to better approximate the probability, we adopt KL-divergence minimization between the two introduced distributions to preserve relationships among different pairwise similarities.

Our main contributions include:
\begin{itemize}
  \item We investigate the online hashing problem by modeling the similarity distribution, instead of only exploiting the pairwise similarities that suffer from a poor generalization problem. The Gaussian normalization is introduced to smooth the extremely imbalanced distribution, while a scaling $t$-Student distribution is proposed to solve the initialization problem, and bridge the gap between the known and unknown distributions.

  \item We propose to align the distributions via KL-divergence between the input data and the binary space, which imposes an intuitive similarity constraint to update hash functions on the new streaming data with a powerful generalizing ability to the past data.

  \item Extensive experiments conducted on three widely-used benchmarks, \emph{i.e.}, CIFAR$10$, Places$205$ and MNIST, demonstrate that the proposed SDOH achieves the best performance over the state-of-the-art methods \cite{huang2013online,cakir2015adaptive,leng2015online,cakir2017online,fatih2017mihash,lin2018supervised,lin2019towards}.
\end{itemize}

\section{Related Work}
According to the learning type, online hashing can be categorized into the SGD-based methods and matrix sketch-based methods.

SGD-based online hashing employs SGD to update the learned parameters.
To our best knowledge, Online Kernel Hashing (OKH) \cite{huang2013online} is the first of this kind, which requires pairs of points to update the hash functions via an online passive-aggressive strategy \cite{crammer2006online}.
Adaptive Hashing (AdaptHash) \cite{cakir2015adaptive} defines a hinge-like loss, which is approximated by a differentiable Sigmoid function adopted to update the hash functions with SGD.
In \cite{cakir2017online}, a more general two-step hashing was introduced, in which binary Error Correcting Output Codes (ECOC) are first assigned to labeled data, and then the hash functions are learned to fit the binary ECOC using Online Boosting.
Cakir \emph{et al}. \cite{fatih2017mihash} developed an Online Hashing with Mutual Information (MIHash), which targets at optimizing the mutual information between the neighbors and non-neighbors, given a query.
Lin \emph{et al}. \cite{lin2018supervised,lin2019hadamard} proposed a Hadamard Codebook based Online Hashing (HCOH), where a more discriminative Hadamard matrix is used as the ECOC codebook to guide the learning of hash functions.

The inspiration of matrix sketch-based online hashing methods comes from the idea of ``data sketch" \cite{liberty2013simple}, where a small size of sketch data is leveraged to preserve the main property of a large-scale dataset. %
To this end, Leng \emph{et al}. \cite{leng2015online} proposed an Online Sketching Hashing (SketchHash), which employs an efficient variant of SVD decomposition to learn hash functions, with a PCA-based batch learning on the sketch to learn hashing weights.
A faster version of Online Sketch Hashing (FROSH) was developed in \cite{Chen2017FROSHFO}, where the independent Subsampled Randomized Hadamard Transform (SRHT) is employed on different data chunks to make the sketch more compact and accurate, and to accelerate the sketching process.

However, existing sketch-based online hashing methods are unsupervised, which suffer from a low performance due to the lack of supervised labels.
SGD-based methods \cite{huang2013online,cakir2015adaptive,cakir2017online,fatih2017mihash,lin2018supervised,lin2019towards} aim to make full use of labels, which still face practical problems as discussed in Sec.\,\ref{introduction}. For OKH \cite{huang2013online}, AdaptHash \cite{cakir2015adaptive}, MIHash \cite{fatih2017mihash} and BSODH \cite{lin2019towards}, less generalization ability exists since only pairwise relationships of current sequential data are considered. As for OSH \cite{cakir2017online} and HCOH \cite{lin2018supervised,lin2019hadamard}, a well-defined ECOC codebook has to be given in advance, which still fails when the size of codebook is inconsistent with the length of hashing bits.

\section{Proposed Method}
\subsection{Problem Definition}
Suppose there is a dataset $\mathbf{X} = [\mathbf{x}_1, ..., \mathbf{x}_n] \in \mathbb{R}^{d \times n}$ with its corresponding labels $\mathbf{L} =  [l_1, ..., l_n] \in \mathbb{N}^{n}$, where $\mathbf{x}_i \in \mathbb{R}^d$ is the $i$-th instance with its class label $l_i \in \mathbb{N}$.
Assume there are $k$ hash functions to be learned, which map each $\mathbf{x}_i \in \mathbb{R}^d$ into a $k$-bit binary code $\mathbf{b}_n = [b_{n1}, ..., b_{nk}]^T \in \{-1, +1\}^{k \times 1}$, and the $k$-th binary bit $b_{nk}$ of $\mathbf{x}_n$ is computed as follows:
\begin{equation}
    b_{nk} = f_k(\mathbf{x}_n) = sgn(\mathbf{w}_k^T\mathbf{x}_n),
\end{equation}
where $f_k$ is the $k$-th hash function, and $\mathbf{w}_k \in \mathbb{R}^d$ is a projection of $f_k$. The $sgn(u)$ function returns $1$, if $u > 0$, and $-1$ otherwise.

Let $\mathbf{W} = [\mathbf{w}_1, ..., \mathbf{w}_k] \in \mathbb{R}^{d \times k}$ be the projection matrix. Then, the binary codes of $\mathbf{X}$ can be computed as:
\begin{equation} \label{sgn}
    \mathbf{B} = [\mathbf{b}_1,...,\mathbf{b}_n] = sgn(\mathbf{W}^T\mathbf{X}).
\end{equation}

Online hashing aims to resolve an online retrieval task by updating hash functions from a sequence of data instances one at a time. Therefore, $\mathbf{X}$ is not available once for all.
Without loss of generality, we denote $\mathbf{X}^t = [\mathbf{x}_{1}^t, ..., \mathbf{x}^t_{n_t} ]^T \in \mathbb{R}^{d \times n_t}$ as the input streaming data at the $t$-stage, $\mathbf{B}^t = [\mathbf{b}^t_1, ..., \mathbf{b}^t_n] \in \{-1, +1\}^{k \times n_t}$ as the learned binary codes for $\mathbf{X}^t$ and $\mathbf{L}^t = [l_{1}^t, ..., l_{n_t}^t] \in \mathbb{N}^{n_t}$ as the corresponding label set, where $n_t$ is the size of data at the $t$-stage.
Further, we denote $\mathbf{S}^t \in \mathbb{R}^{n_t \times n_t}$ as the pairwise similarity matrix, where $\mathbf{S}^t_{ij} = 1$, if $\mathbf{x}_i^t$ and $\mathbf{x}_j^t$ share the same label, otherwise $\mathbf{S}^t_{ij} = 0$.
In an online setting, the parameter matrix $\mathbf{W}^t$ should be updated based on the current data $\mathbf{X}^t$ only.

\subsection{Proposed Framework \label{scheme}}
The framework of the proposed method can be seen in Fig.\,\ref{system}.
Specifically, suppose that at the $t$-th round, we have a known similarity distribution matrix $\mathcal{P}^t \in \mathbb{R}^{n_t \times n_t}$ and an unknown Hamming distance distribution matrix $\mathcal{Q}^t \in \mathbb{R}^{n_t \times n_t}$.
The goal of the proposed SDOH is to align $\mathcal{Q}^t$ with $\mathcal{P}^t$, such that the similarity can be well preserved in the Hamming space. It is achieved by minimizing the KL-divergence as follows:
\begin{equation} \label{kl}
    KL(\mathcal{P}^t||\mathcal{Q}^t) = \sum_{ij}\mathcal{P}_{ij}^t\log\frac{\mathcal{P}_{ij}^t}{\mathcal{Q}_{ij}^t},
\end{equation}
where $\mathcal{P}^t_{ij}$ and $\mathcal{Q}^t_{ij}$ are the $j$-th elements in the $i$-th row of $\mathcal{P}^t$ and $\mathcal{Q}^t$, respectively. In the following, we elaborate on the details of $\mathcal{P}^t$ and $\mathcal{Q}^t$.

%
%In the following, we elaborate the imbalanced distribution problem and poor initialization problem for existing $\mathcal{P}^t$ and $\mathcal{Q}^t$, respectively.
%
%And then, new and well defined formulations for $\mathcal{P}^t$ and $\mathcal{Q}^t$ are introduced to well address these issues.

\subsubsection{Gaussian-based Normalization}
One common approach to obtain $\mathcal{P}^t$ is to normalize the similarity matrix $\mathbf{S}^t$ with each element $\mathcal{P}^t_{ij}$ as:
\begin{equation} \label{ori_p}
        \mathcal{P}^t_{ij} = \frac{\mathbf{S}^t_{ij}}{\sum_{i \neq j}\mathbf{S}^t_{ij}}.
\end{equation}

However, such a probability matrix may suffer from an extremely imbalanced distribution, as shown in Fig.\,\ref{system}(a).
For instance, when $\mathbf{S}^t$ is a highly sparse matrix that is common in an online setting \cite{lin2019towards}, $\mathcal{P}^t_{ij}$ is with a higher probability if $\mathbf{S}^t_{ij} = 1$ and $\mathcal{P}^t_{ij}log\frac{\mathcal{P}^t_{ij}}{\mathcal{Q}^t_{ij}}$ grows quickly.
Similarly, if $\mathbf{S}^t_{ij} = 0$, $\mathcal{P}^t_{ij}log\frac{\mathcal{P}^t_{ij}}{\mathcal{Q}^t_{ij}}$ decreases to 0 quickly.

Therefore, the learning of Eq.(\,\ref{ori_p}) heavily relies on similar pairs and thus loses the information of dissimilar pairs, as shown in Fig.\,\ref{system}(c).
To address this issue, one key novelty in our proposed SDOH is to modify Eq.(\,\ref{ori_p}) as:
\begin{equation} \label{mod_p}
        \mathcal{P}^t_{ij} = \frac{f(\mathbf{S}^t_{ij})}{\sum_{i \neq j}f(\mathbf{S}^t_{ij})},
\end{equation}
where $f(\cdot)$ is introduced to smooth the imbalanced distribution as shown in Fig.\,\ref{system}(b).
We assume that $\mathbf{S}^t_{ij}$ follows a Gaussian distribution widely used in practice, \emph{i.e.}, $\mathbf{S}^t_{ij} \sim N({\mu}, {{\sigma}}^2)$, where $\mu$ and $\sigma$ are the mean and variance of the pairwise similarity distribution, respectively. Therefore, we derive $f(\mathbf{S}^t_{ij}) = \frac{1}{\sqrt{2\pi}{\sigma}}$exp$(-\frac{(\mathbf{S}^t_{ij} - {\mu})^2}{2{{\sigma}}^2})$, where exp$(\cdot)$ is the exponential function.
Different values of the pair $({\mu}, {\sigma})$ have different impacts on $\mathcal{P}^t_{ij}$. To sum up, $\mu$ decides the position of the highest value of $f(\cdot)$. The larger the $\sigma$ is, the smoother the function $f(\cdot)$ is. Therefore, Eq.(\,\ref{mod_p}) can well alleviate the imbalanced distribution problem caused by Eq.(\,\ref{ori_p}) (see Fig.\,\ref{system}(d)).

\subsubsection{Scaling Student $t$-distribution}
For the distribution $\mathcal{Q}^t$, we define it as the probability of Hamming distance.
The similarity between $b_i$ and $b_j$ can be measured by the Hamming distance as:
\begin{equation}\label{dis_ham}
    dist(\mathbf{b}^t_i, \mathbf{b}^t_j) = \frac{1}{4}{\|\mathbf{b}^t_i - \mathbf{b}^t_j \|_2^2}.
\end{equation}

We propose a scaling Student $t$-distribution based on a new $\mathcal{Q}^t$ with one degree of freedom to transform Hamming distances into probabilities.
We start from the works in \cite{maaten2008visualizing,lin2015semantics}, and each element of the original $\mathcal{Q}^t_{ij}$ is defined as:
\begin{equation} \label{ori_q}
    \mathcal{Q}^t_{ij} = \frac{(1 + dist(\mathbf{b}^t_i, \mathbf{b}^t_j))^{-1}}{\sum_{k \neq m}(1 + dist(\mathbf{b}^t_k, \mathbf{b}^t_m))^{-1}}.
\end{equation}

However, such an assigned distribution causes an unsatisfactory initialization of $\mathcal{Q}^t$, which may lead to the performance degradation. Ideally, if $\mathbf{S}^t_{ij} = 1$, we need a higher value of $\mathcal{Q}^t_{ij}$.
However, the value of $\mathcal{Q}_{ij}^t$ in Eq.(\,\ref{ori_q}) depends on the initialization of $\mathbf{W}^t$.
If $\mathbf{W}^t$ does not initialize well, $\mathcal{Q}^t_{ij}$ is likely to be very small for $\mathbf{S}^t_{ij} = 1$. In such a case, $\mathcal{P}^t_{ij}\log\frac{\mathcal{P}^t_{ij}}{\mathcal{Q}^t_{ij}}$ in Eq.(\,\ref{kl}) grows quickly. Similarly, when $\mathbf{S}^t_{ij} = 0$, $\mathcal{P}^t_{ij}\log\frac{\mathcal{P}^t_{ij}}{\mathcal{Q}^t_{ij}}$ in Eq.(\,\ref{kl}) may be very small.
Therefore, Eq.(\,\ref{ori_q}) may result in an extremely poor initialization (see Fig.\,\ref{system}(e)).

To avoid such a poor initialization, another key novelty in our approach is to revise Eq.(\,\ref{ori_q}) as follows:
\begin{equation} \label{mod_q}
    \mathcal{Q}^t_{ij} = \frac{(1 + \frac{dist(\mathbf{b}^t_i, \mathbf{b}^t_j)}{{\eta}_{ij}})^{-1}}{\sum_{k \neq m}(1 + \frac{dist(\mathbf{b}^t_k, \mathbf{b}^t_m)}{{\eta}_{km}})^{-1}},
\end{equation}
where the scaling parameter ${\eta}_{ij} = p$ if $\mathbf{S}^t_{ij} = 1$, otherwise ${\eta}_{ij} = n$.
To analyze, scaling up ${\eta}_{ij}$ will increase the value of $\mathcal{Q}^t_{ij}$, thus further decrease the value of $\mathcal{P}^t_{ij}\log\frac{\mathcal{P}^t_{ij}}{\mathcal{Q}^t_{ij}}$ when $\mathbf{S}^t_{ij} = 1$.
Similar analyses can be applied to the case of $\mathbf{S}^t_{ij} = 0$ (see Fig.\,\ref{system}(f)). Therefore, Eq.(\,\ref{mod_q}) can well reduce the influence of initialization problem (see Fig.\,\ref{system}(d)).

The final objective function can be derived:
\begin{equation}\label{loss}
    \mathcal{L}^t(\mathbf{W}^t) = \min_{\mathbf{W}^t}\sum_{ij}\mathcal{P}_{ij}^t\log\frac{\mathcal{P}_{ij}^t}{\mathcal{Q}_{ij}^t}.
\end{equation}

We note that the proposed SDOH clearly differs from SePH \cite{lin2015semantics} as:
$1)$ Our SDOH is based on a new and well defined $\mathcal{P}^t$ and $\mathcal{Q}^t$ which enable to avoid the imbalanced distribution and poor initialization problems.
$2)$ We are the first to employ $t$-Student distribution for online uni-modal retrieval, while SePH aims at solving offline cross-modal retrieval.
$3)$ Our SDOH is implemented in an end-to-end manner, where the learning of hash functions is integrated into KL-divergence. While SePH is based on a two-step framework, where KL-divergence is used to guide the learning of binary codes first, and then hash functions are learned to approximate the learned binary codes.

\subsection{The Optimization}
After obtaining the distributions $\mathcal{P}^t$ and $\mathcal{Q}^t$, we aim to optimize the KL-divergence in Eq.(\,\ref{loss}) to preserve the similarities in the Hamming space.
Due to the discrete sign function in Eq.(\,\ref{sgn}), the above objective function is highly non-convex and difficult (usually NP-hard) to optimize.
%
%A feasible solution is to relax the discrete constraint by omitting the sign function, which is widely applied to the hashing code learning.
%
%However, this approximation may accumulate considerable quantization errors, which makes the final binary codes less effective.
%
To solve it, we follow the work in \cite{liu2017ordinal} to replace the sign function with the tanh function as follows:
\begin{equation} \label{tanh}
    \mathbf{B}^t = [{\mathbf{b}_1^t},...,{\mathbf{b}_{n_t}^t}] = tanh\big({(\mathbf{W}^t)}^T\mathbf{X}^t\big).
\end{equation}

To solve the optimization problem in Eq.(\,\ref{loss}), we adopt the SGD algorithm to update the hash functions at the $t$-stage as below:
\begin{equation} \label{update}
    \mathbf{W}^{t+1} \leftarrow \mathbf{W}^t - {\lambda}\frac{\partial \mathcal{L}^t(\mathbf{W}^t)}{\partial \mathbf{W}^t},
\end{equation}
where ${\lambda}$ is a positive learning rate.

We now elaborate the partial derivative of $\mathcal{L}$ \emph{w.r.t.} $\mathbf{W}^t$, \emph{i.e.}, $\frac{\partial \mathcal{L}^t(\mathbf{W}^t)}{\partial \mathbf{W}^t}$. The gradient \emph{w.r.t.} $\mathbf{b}^t_i$ can be computed as follows:
\begin{equation} \label{par_b}
    \frac{\partial \mathcal{L}^t(\mathbf{B}^t)}{\partial \mathbf{b}_i^t} = \sum_{j \neq i}{(\mathcal{P}^t_{ij} - \mathcal{Q}^t_{ij})(1 + \frac{d(\mathbf{b}_i^t, \mathbf{b}_j^t)}{{\eta}_{ij}})^{-1}(\mathbf{b}^t_i - \mathbf{b}^t_j)}.
\end{equation}

Further, we denote $\mathbf{T}^t = (1 + \mathbf{D}^t) \in \mathbb{R}^{n_t \times n_t}$, where $\mathbf{D}^t$ is the matrix with $\mathbf{D}^t_{ij}$ being $\frac{d(\mathbf{b}_i^t, \mathbf{b}_j^t)}{{\eta}_{ij}}$.
Let $\mathbf{L}^t = (\mathcal{P}^t - \mathcal{Q}^t) \odot \mathbf{T}^t \in \mathbb{R}^{n_t \times n_t}$, where $\odot$ stands for the Hadamard product.
Let $\hat{\mathbf{L}}^t = diag(\mathbf{L}^t\mathbf{1}) \in \mathbb{R}^{n_t \times n_t}$, where $diag(\cdot)$ is the diagonal matrix, and $\mathbf{1} \in \mathbb{R}^{n_t}$ represents a vector with all elements being $1$.
Therefore, we obtain the gradient \emph{w.r.t.} $\mathbf{B}^t$ as:
\begin{equation}\label{part_B}
    \frac{\partial \mathcal{L}^t(\mathbf{B}^t)}{\partial \mathbf{B}^t} = \mathbf{B}^t(\hat{\mathbf{L}}^t - \mathbf{L}^t).
\end{equation}
Besides, it is easy to obtain the derivative of $\mathbf{B}^t$ \emph{w.r.t.} $\mathbf{W}^t$ based on Eq.(\,\ref{tanh}) as follows:
\begin{equation} \label{part_BW}
    \frac{\partial \mathbf{B}^t}{\partial \mathbf{W}^t} = \mathbf{X}^t({\mathbf{P}^t})^T,
\end{equation}
where $\mathbf{P}^t = 1 - tanh\big(({\mathbf{W}^t)}^T\mathbf{X}^t\big) \odot tanh\big(({\mathbf{W}^t)}^T\mathbf{X}^t\big) \in \mathbb{R}^{k\times n_t}$.
Combining Eq.(\,\ref{part_B}) and Eq.(\,\ref{part_BW}), we obtain the derivative of $\mathcal{L}$ \emph{w.r.t.} $\mathbf{W}^t$ via the chain rule as:
\begin{equation} \label{part_W}
    \frac{\partial \mathcal{L}^t(\mathbf{W}^t)}{\partial \mathbf{W}^t} = \mathbf{X}^t(\hat{\mathbf{L}}^t - \mathbf{L}^t)(\mathbf{B}^t \odot {\mathbf{P}^t})^T.
\end{equation}

The optimization process for the proposed SDOH is summarized in the supplement material with more details.

%%%%%%%%%%%%%%%%%%%%%%%%%%%%%%%%%%%%%%%%%%%%%%%%%%%%%%%%%%%%%%%%%%%%%%%%%%%
\begin{table*}[]
\centering
\begin{tabular}{c|cccc|cccc}
\hline
\multirow{2}{*}{Method} & \multicolumn{4}{c|}{\textit{m}AP}                   & \multicolumn{4}{c}{Precision@H2}          \\
\cline{2-9}
                        & 32-bit &48-bit & 64-bit & 128-bit & 32-bit &48-bit   & 64-bit  &128-bit \\
\hline
OKH                     &0.223   &0.252  &0.268   &0.350    &0.100   &0.452    & 0.175   &0.372  \\
\hline
SketchHash              &0.302   &0.327  &  -     &   -     &0.385   &0.059    &   -     & -     \\
\hline
AdaptHash               &0.216   &0.297  &0.305   &0.293    &0.185   &0.093    &0.166    &0.164         \\
\hline
OSH                     &0.129   &0.131  &0.127   &0.125    &0.137   &0.117    &0.083    &0.038  \\
\hline
MIHash                  &0.675   &0.668  &0.667   &0.664    &0.657   &0.604    &0.500    &0.413    \\
\hline
HCOH                    &0.688   &\underline{0.707}&\underline{0.724}&\underline{0.734}&\underline{0.731}
                        &0.694&0.633&0.471  \\
\hline
BSODH                   &\underline{0.689}&0.656  &0.709   &0.711    &0.691&\underline{0.697}&\underline{0.690}&\textbf{0.602}  \\
\hline
\hline
SDOH                    &\textbf{0.765}&\textbf{0.762}&\textbf{0.751}&\textbf{0.742}&\textbf{0.785}&\textbf{0.781}
                        &\textbf{0.734}    &\underline{0.550}\\
\hline
\end{tabular}
\vspace{0.2em}
\caption{\textit{m}AP and Precision@H2 comparisons on CIFAR-$10$ with $32$, $48$, $64$ and $128$ bits.}
\label{cifar-map-precisionh2}
\vspace{-1em}
\end{table*}

%%%%%%%%%%%%%%%%%%%%%%%%%%%%%%%%%%%%%%%%%%%%%%%%%%%%%%%%%%%%%%%%%%
\begin{figure*}[!t]
\begin{center}
\includegraphics[height=0.24\linewidth]{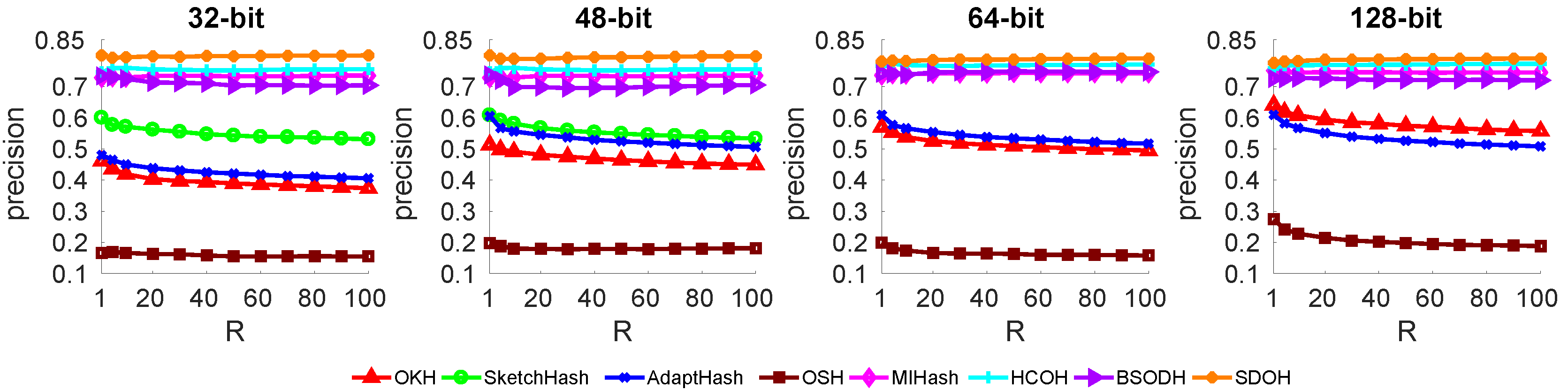}
\caption{\label{cifar_precision} Precision@R curves of compared algorithms on CIFAR-$10$.}
\end{center}
\vspace{-1.3em}
\end{figure*}
%%%%%%%%%%%%%%%%%%%%%%%%%%%%%%%%%%%%%%%%%%%%%%%%%%%%%%%%%%%%%%%%%%
\begin{figure*}[!t]
\begin{center}
\includegraphics[height=0.24\linewidth]{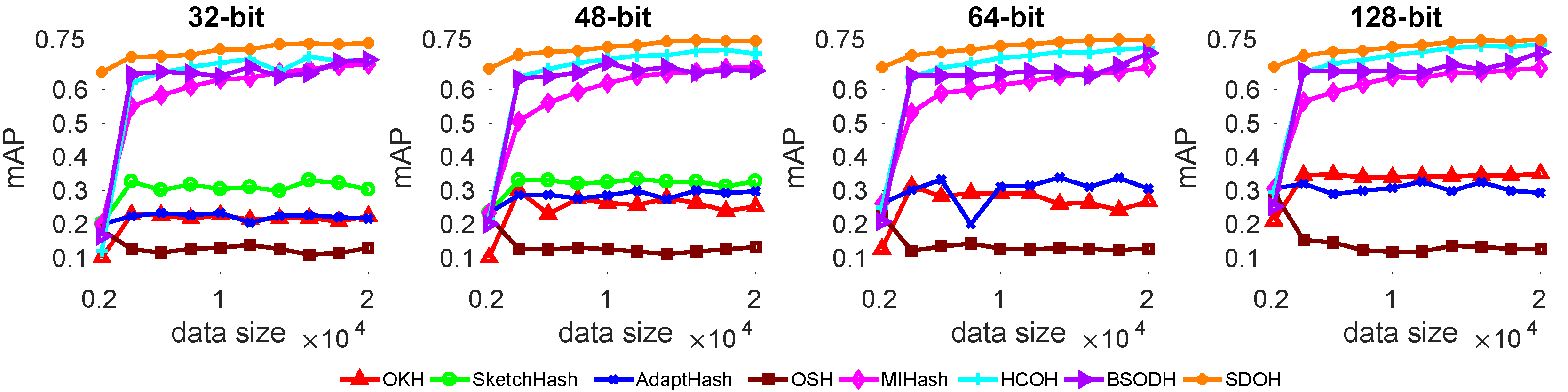}
\caption{\label{cifar_map} \textit{m}AP performance with respect to different sizes of training instances on CIFAR-$10$.}
\end{center}
\vspace{-1.8em}
\end{figure*}

%%%%%%%%%%%%%%%%%%%%%%%%%%%%%%%%
\begin{figure}
\begin{center}
\centerline{
\includegraphics[height=0.33\linewidth]{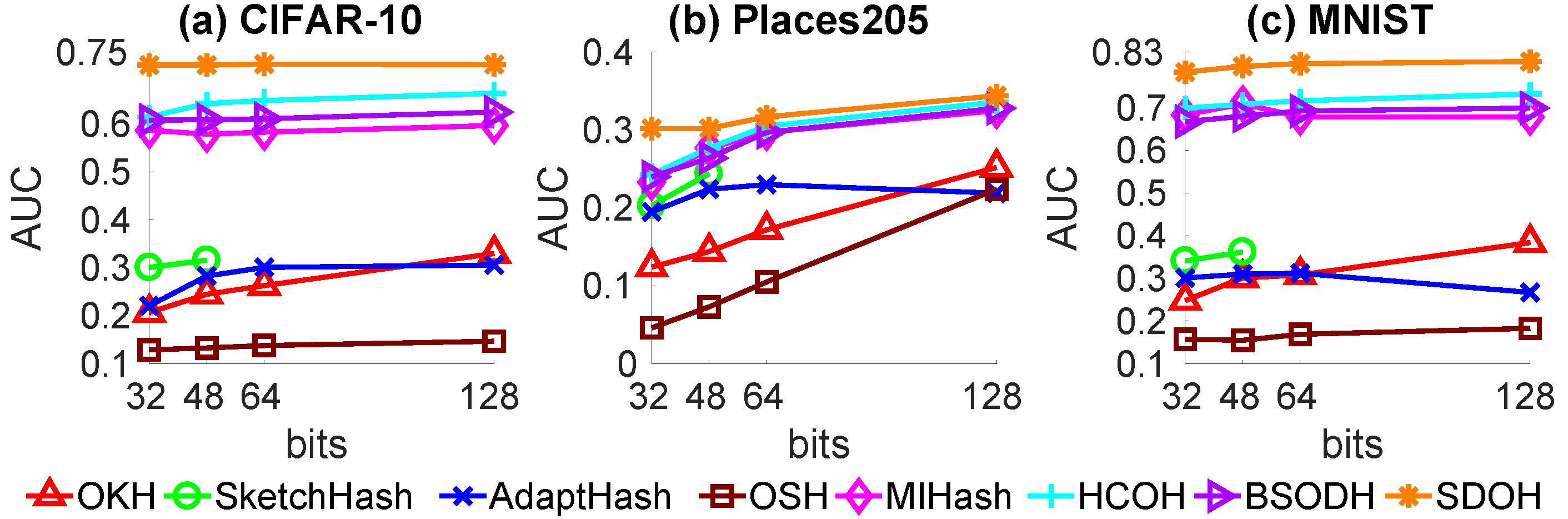}}
\caption{\label{auc}AUC curves of compared algorithms on three datasets with the hashing bits of different lengths.}
\end{center}
\vspace{-3.8em}
\end{figure}
%%%%%%%%%%%%%%%%%%%%%%%%%%%%%%%%%%%%%%%%%%%%%%%%%%%%%%%%%%%%%%%%%%

\section{Experiments \label{experiment}}
We conduct experiments on three benchmark datasets, \textit{i.e.}, CIFAR-$10$ \cite{krizhevsky2009learning}, Places$205$ \cite{zhou2014learning}, and MNIST \cite{lecun1998gradient}.
Our proposed SDOH will be compared with several state-of-the-art online hashing methods \cite{huang2013online,cakir2015adaptive,leng2015online,cakir2017online,fatih2017mihash,lin2018supervised,lin2019towards} to demonstrate its performance.

\subsection{Experimental Settings}

\textbf{Datasets.} The CIFAR-$10$ dataset consists of $60$K instances from $10$ categories.
Each instance is represented by a $4,096$-dim CNN feature vector, extracted from VGG-$16$ \cite{simonyan2014very}.
Similar to \cite{fatih2017mihash}, we randomly split the dataset into a retrieval set with $59$K samples, and a test set with $1$K samples.
Besides, $20$K training images from the retrieval set are sampled to learn the hash functions.

The Places$205$ dataset is a large-scale and challenging dataset contains more than $2.5$ million images with $205$ scenes.
We extract CNN features from the $fc7$ layer of AlexNet \cite{krizhevsky2012imagenet}, which are reduced into $128$-dim features by PCA.
Following \cite{cakir2017online}, $20$ images from each scene are used to construct a test set, and the remaining is used as the retrieval set.
A random subset of $100$K images is used to update the hash functions.

The MNIST dataset consists of $70$K handwritten digit images with $10$ classes.
Following \cite{lin2018supervised}, each image is represented by $28 \times 28 = 784$-dim normalized pixel values.
The test set is constructed by sampling $100$ instances from each class, and the others are used to form the retrieval set. %
Besides, $20$K images from the retrieval set are sampled to learn the hash functions.

%%%%%%%%%%%%%%%%%%%%%%%%%%%%%%%%%%%%%%%%%%%%%%
\begin{table*}[]
\centering
\begin{tabular}{c|cccc|cccc}
\hline
\multirow{2}{*}{Method} & \multicolumn{4}{c|}{\textit{m}AP@$1,000$}                   & \multicolumn{4}{c}{Precision@H2}          \\
\cline{2-9}
                        & 32-bit &48-bit  &64-bit & 128-bit & 32-bit &48-bit   & 64-bit  & 128-bit \\
\hline
OKH                     &0.122   &0.048   &0.114  &0.258    &0.026   &0.017    &0.217    &0.075    \\
\hline
SketchHash              &0.202   &0.242   &  -    &  -      &0.220   &0.176    &  -      &-   \\
\hline
AdaptHash               &0.195   &0.223   &0.222  &0.229    &0.012   &0.185    &0.021    &0.022      \\
\hline
OSH                     &0.022   &0.032   &0.043  &0.164    &0.012   &0.023    &0.030    &0.059     \\
\hline
MIHash                  &0.244   &\underline{0.288}   &0.308  &0.332    &0.204  &0.242   &0.202&0.069   \\
\hline
HCOH                    &\underline{0.259}&0.280   &\underline{0.321}  &\underline{0.347}    &\textbf{0.252}   &0.179    &0.114    &0.036     \\
\hline
BSODH                   &0.250   &0.273   &0.308  &0.337&0.241&\underline{0.246}&\underline{0.212}&\underline{0.101}  \\
\hline
\hline
SDOH                    &\textbf{0.306}&\textbf{0.309}&\textbf{0.324}&\textbf{0.348}&\underline{0.249}&\textbf{0.248}&\textbf{0.217}&\textbf{0.103}  \\
\hline
\end{tabular}
\vspace{0.2em}
\caption{\textit{m}AP@$1,000$ and Precision@H2 comparisons on Places$205$ with $32$, $48$, $64$ and $128$ bits.}
\label{places-map-precisionh2}
\vspace{-1em}
\end{table*}

%%%%%%%%%%%%%%%%%%%%%%%%%%%%%%%%%%%%%%%%%%%%%%%%%%%%%%%%%%%%%%%%%%
\begin{figure*}[!t]
\begin{center}
\includegraphics[height=0.24\linewidth]{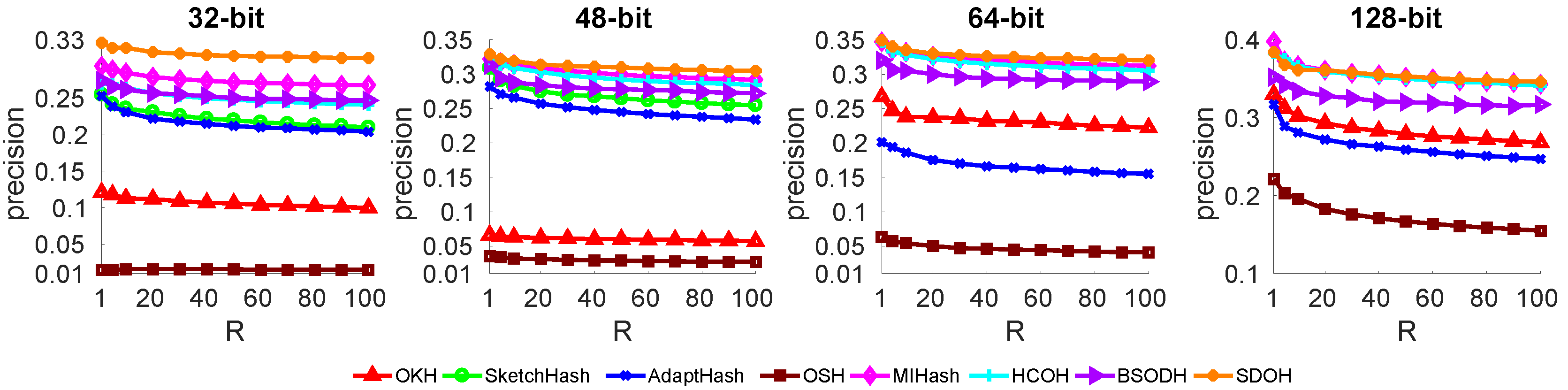}
\caption{\label{places_precision} Precision@R curves of compared algorithms on Places$205$.}
\end{center}
\vspace{-1.3em}
\end{figure*}
%%%%%%%%%%%%%%%%%%%%%%%%%%%%%%%%%%%%%%%%%%%%%%%%%%%%%%%%%%%%%%%%%%
\begin{figure*}[!t]
\begin{center}
\includegraphics[height=0.24\linewidth]{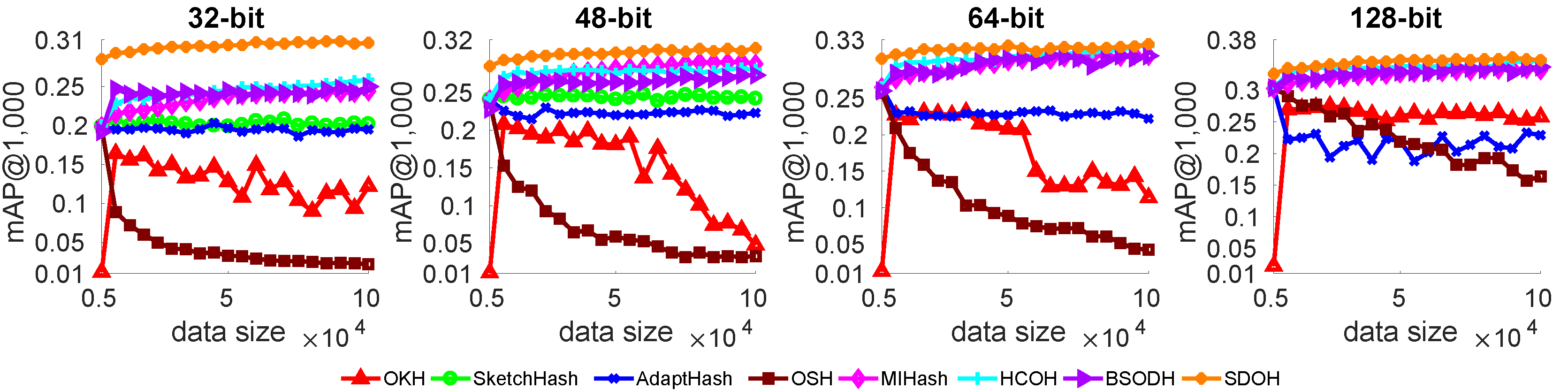}
\caption{\label{places_map} \textit{m}AP performance with respect to different sizes of training instances on Places$205$.}
\end{center}
\vspace{-2.5em}
\end{figure*}

\textbf{Compared Methods.} The proposed SDOH is compared with seven state-of-the-art online hashing methods, including the Online Kernel Hashing ($\mathbf{OKH}$) \cite{huang2013online}, Online Sketch Hashing ($\mathbf{SketchHash}$) \cite{leng2015online}, Adaptive Hashing ($\mathbf{AdaptHash}$) \cite{cakir2015adaptive}, Online Supervised Hashing ($\mathbf{OSH}$) \cite{cakir2017online}, Online hashing with Mutual Information ($\mathbf{MIHash}$) \cite{fatih2017mihash}, Hadamard Codebook based Online Hashing ($\mathbf{HCOH}$) \cite{lin2018supervised} and Balanced Similarity for Online Discrete Hashing (BSODH) \cite{lin2019towards}.
Our model is implemented with MATLAB.
The training is done on a standard workstation with a $3.6$GHz Intel Core I$7$ $4790$ CPU and $16$G RAM.
The source codes of these methods are publicly available.
We carefully follow the original parameter settings for each method and report their best results.

%%%
\textbf{Evaluation Protocols.} We use five widely-adopted evaluation metrics for performance comparisons, including mean Average Precision (denoted as \textit{mAP}), precision within a Hamming ball of radius $2$ centered on each query (denoted as \textit{Precision@H$2$}), mean precision of the top R retrieved neighbors (denoted as \textit{Precision@R}), \textit{m}AP vs. different sizes of training instances, as well as its corresponding area under the \textit{m}AP curve (denoted as \textit{AUC}).
Noticeably, due to the large scale of Places$205$ dataset, it is very time-consuming to compute \textit{m}AP.
Following \cite{fatih2017mihash,lin2018supervised}, we only compute \textit{m}AP on the top $1,000$ retrieved items (denoted as \textit{\textit{m}AP@$1,000$}).
And for SketchHash \cite{leng2015online}, the batch size has to be larger than the size of hashing bits.
Thus, we only report its performance for hashing bits of $32$ and $48$.

\subsection{Results and Discussions}
First, we report the \emph{m}AP (\emph{m}AP@$1,000$) and Precision@H$2$ performance in Tabs.\,\ref{cifar-map-precisionh2}, \ref{places-map-precisionh2} and \ref{mnist-map-precisionh2}. The highest values are shown in boldface, and the second best are with underlines. It can be seen that the proposed SDOH is the best in almost all cases. Interestingly, as the number of hashing bit increases, the proposed SDOH outperforms others by large margins.

\begin{figure*}[!t]
\begin{center}
\includegraphics[height=0.24\linewidth]{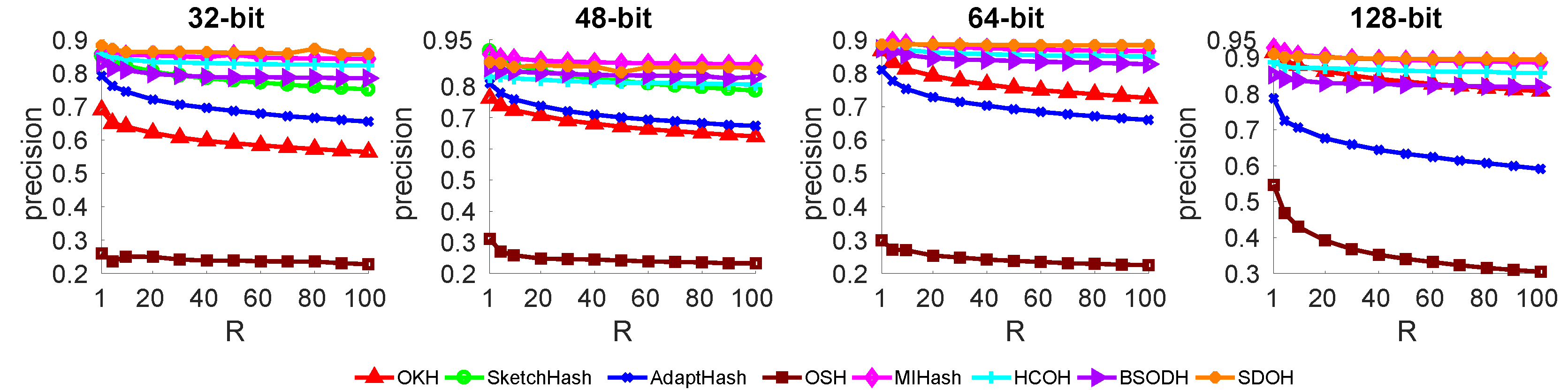}
\caption{\label{mnist_precision} Precision@R curves of compared algorithms on MNIST.}
\end{center}
\vspace{-2em}
\end{figure*}

Second, we analyze the Precision@R performance with R ranging from $1$ to $100$ on the three benchmarks in Figs.\,\ref{cifar_precision}, \ref{places_precision} and \ref{mnist_precision}. SDOH achieves super results on all three benchmarks in all hashing bits, which demonstrates the excellent performance of the proposed method.

\begin{figure*}[!t]
\begin{center}
\includegraphics[height=0.24\linewidth]{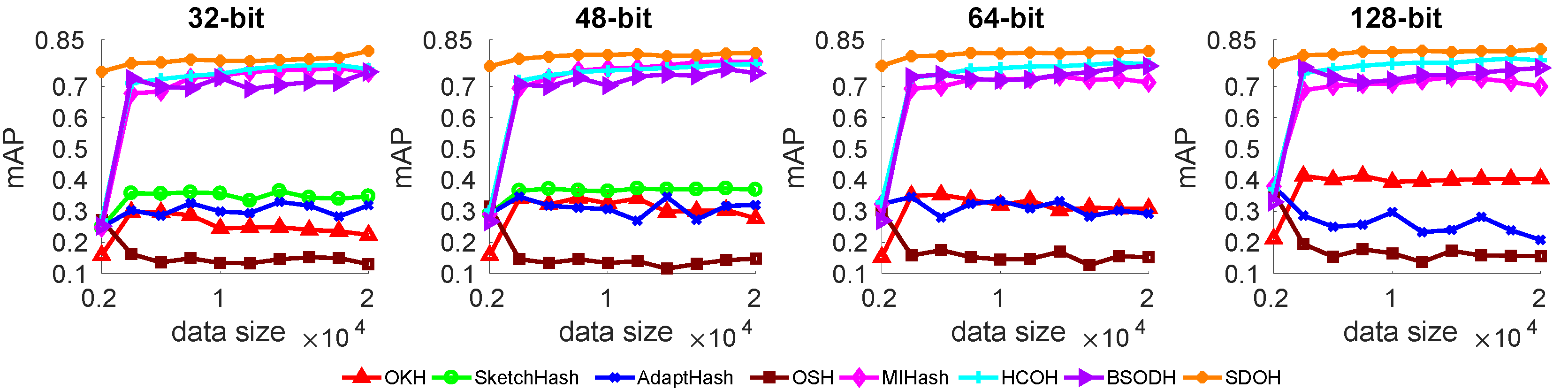}
\caption{\label{mnist_map} \textit{m}AP performance with respect to different sizes of training instances on MNIST.}
\end{center}
\vspace{-2.2em}
\end{figure*}

Third, we report the \emph{m}AP (\emph{m}AP@$1,000$) measure \emph{w.r.t.} different training sizes in Figs.\,\ref{cifar_map}, \ref{places_map} and \ref{mnist_map}.
As the size of training data increases, SDOH has consistently higher \emph{m}AP (\emph{m}AP@$1,000$) on all three benchmarks.
To quantitatively evaluate the performance of all methods, we take a deeper analysis in terms of their AUC metrics in Fig.\,\ref{auc}.
For CIFAR-$10$ and MNIST, the AUC performance of Fig.\,\ref{cifar_map} and Fig.\,\ref{mnist_map} is charted in Fig.\,\ref{auc}(a) and Fig.\,\ref{auc}(c), respectively. Obviously, in all hashing bits, SDOH outperforms other methods by a large margin.

%%%%%%%%%%%%%%%%%%%%%%%%%%%%%%%%%%%%%%%%%%%%%%%%%%%%%%%%%%
\begin{table*}[]
\centering
\begin{tabular}{c|cccc|cccc}
\hline
\multirow{2}{*}{Method} & \multicolumn{4}{c|}{\textit{m}AP}      & \multicolumn{4}{c}{Precision@H2}          \\
\cline{2-9}
                        &32-bit  &48-bit & 64-bit & 128-bit & 32-bit & 48-bit & 64-bit & 128-bit \\
\hline
OKH                     &0.224   &0.273  &0.301  &0.404    &0.457   &0.724   &0.522   &0.126    \\
\hline
SketchHash              &0.348   &0.369  &  -    &  -      &0.691   &0.251   & -      &-    \\
\hline
AdaptHash               &0.319   &0.318  &0.292  &0.208    &0.535   &0.335   &0.163   &0.168    \\
\hline
OSH                     &0.130   &0.148  &0.146  &0.143    &0.192   &0.134   &0.109   &0.019    \\
\hline
MIHash                  &0.744   &\underline{0.780}&0.713  &0.681    &0.814   &0.739   &0.720&0.471  \\
\hline
HCOH                    &\underline{0.756}&0.772&0.759&\underline{0.771}&\underline{0.826}&0.766   &0.643   &0.370 \\
\hline
BSODH                   &0.747&0.743&\underline{0.766}   &0.760 &\underline{0.826}&\underline{0.804}&\underline{0.814}   &\underline{0.643}  \\
\hline
\hline
SDOH              &\textbf{0.814} &\textbf{0.799}  &\textbf{0.802}  &\textbf{0.823}   &\textbf{0.835}   &\textbf{0.833}   &\textbf{0.850}   &\textbf{0.828}  \\
\hline
\end{tabular}
\vspace{0.2em}
\caption{\textit{m}AP and Precision@H$2$ comparisons on MNIST with $32$, $48$, $64$ and $128$ bits.}
\label{mnist-map-precisionh2}
\vspace{-1em}
\end{table*}

%%%%%%%%%%%%%%%%%%%%%%%%%%%%%%%%%%%%%%%%%%%%%%%%%%%%%%%%%%%%%%%%%%
%%%%%%%%%%%%%%%%%%%%%%%%%%%%%%%%%%%%%%%%%%%%%%%%%%%%%%%%%%%%%%%%%%

%%%%%%%%%%%%%%%%%%%%%%%%%%%%%%%%
\begin{figure}
\begin{center}
\centerline{
\includegraphics[height=0.34\linewidth]{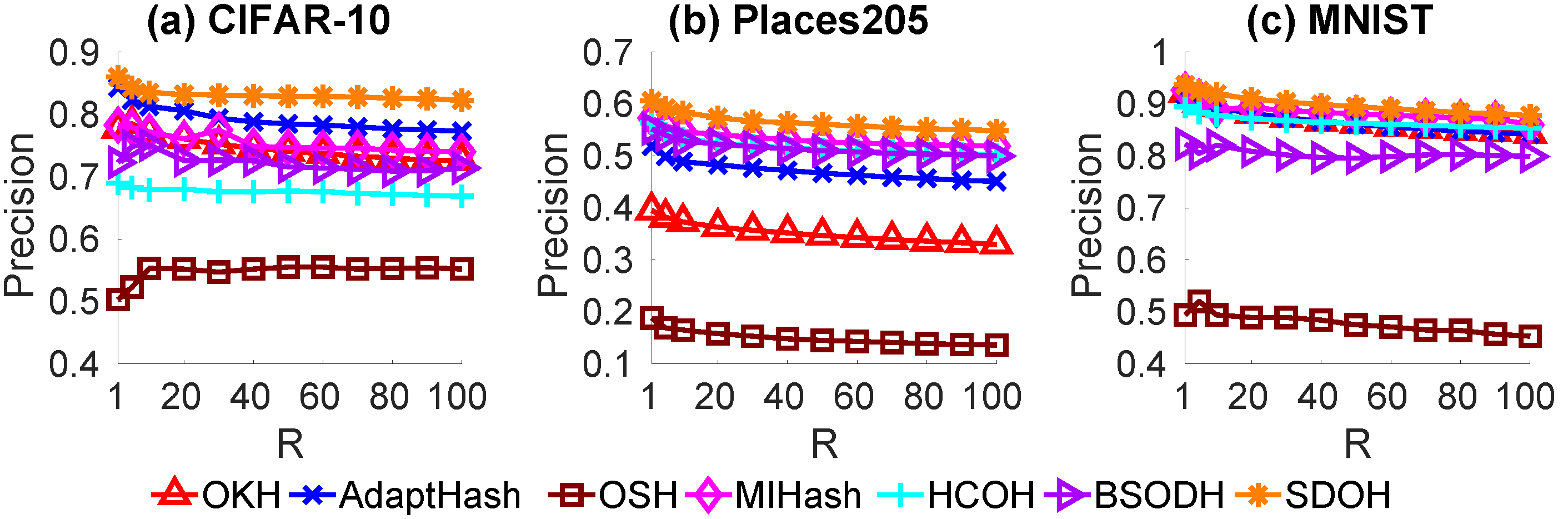}}
\caption{\label{unseen}The Precision@R curves on unseen class when the hash bit is $64$.}
\end{center}
\vspace{-4.5em}
\end{figure}
%%%%%%%%%%%%%%%%%%%%%%%%%%%%%%%%%%%%%%%%%%%%%%%%%%%%%%%%%%%%%%%%%%

%
On Places$205$, the AUC performance in Fig.\,\ref{places_map} is plotted in Fig.\,\ref{auc}(b). When the hashing bit is $32$, the proposed method also transcends all other methods by a certain margin. Though, similar results can be observed for MIHash and HCOH in other hashing bits, the performance of SDOH still ranks the best.
Quantitatively, compared with the second best method, \emph{i.e.}, HCOH, SDOH achieves an average improvement of $14.40\%$, $0.11\%$ and $11.74\%$ on CIFAR-$10$, Places$205$, and MNIST, respectively.

Note that Figs.\,\ref{cifar_map}, \ref{places_map} and \ref{mnist_map} also validate the generalization ability of the proposed SDOH.
Taking the case of hashing bit $=$ $32$ on CIFAR-$10$ for instance (Fig.\,\ref{cifar_map}), when the size of training data is $2,000$, SDOH already achieves a satisfying result of $0.625$ \emph{m}AP compared with other methods, \emph{e.g.}, $0.200$ \emph{m}AP for MIHash and $0.120$ \emph{m}AP for HCOH.
For a more in-depth analysis, MIHash only considers the pairwise similarities while HCOH restricts the length of hashing bit to be consistent with the size of ECOC codebook applied.
It shows that the proposed method not only captures the pairwise similarities of the current data batch, but also the relationships among data batches at different stages, which demonstrates the usefulness of exploiting the distribution of pairwise similarities.

Furthermore, we find the advantage of KL-based solution in comparison with others, such as inner product.
As shown in the above figures and tables, the inner-product-based BSODH shows advantages mainly in Precision@H$2$.
Nevertheless, the proposed KL-based SDOH still outperforms BSODH by a clear margin, which demonstrates the effectiveness and capability of the KL-based solution.

\subsection{Retrieval of Unseen Classes}
We further test the performance on unseen classes as in \cite{sablayrolles2017should}.
To do that, $75\%$ of the categories are treated as seen classes to form the training set.
The remaining $25\%$ categories are regarded as unseen classes, which are divided into a retrieval set and test set to evaluate the hashing model.
For each query, we retrieve the nearest neighbors among the retrieval set and then compute the Precision@R.
The experiments are done when the hashing bit is $64$.
The results are shown in Fig.\,\ref{unseen}.
Clearly, the proposed SDOH achieves the best performance among all methods, which further demonstrates the generalization capability of the proposed framework for online hashing.

\section{Conclusions}
We have presented a novel online hashing method, named SDOH, which aims to align the similarity distributions between the original data space and the hashing space to preserve the semantic relationship well in the Hamming space.
To achieve the goal, we first transform the discrete similarity matrix into a specified probability matrix via a Gaussian-based normalization to solve the imbalanced distribution problem.
Second, to deal with the poor initialization, we have developed a scaling Student $t$-distribution to transform pairwise Hamming distance computation into a probability estimation problem.
Finally, we approximate these two distributions via the KL-divergence to impose an intuitive similarity constraint to update hash functions with a powerful generalization ability.
Experimental results have shown that the proposed SDOH achieves better results than the state-of-the-art methods in comparison.

\section{Acknowledge}
This work is supported by the National Key R\&D Program (No. 2017YFC0113000, and No. 2016YFB1001503), Nature Science Foundation of China (No. U1705262, No. 61772443, and No. 61572410), Post Doctoral Innovative Talent Support Program under Grant BX201600094, China Post-Doctoral Science Foundation under Grant 2017M612134, Scientific Research Project of National Language Committee of China (Grant No. YB135-49), and Nature Science Foundation of Fujian Province, China (No. 2017J01125 and No. 2018J01106).

{\small
\bibliographystyle{ieee}
\bibliography{mybib}
}

\end{document}